\pdfoutput=1
\documentclass[11pt]{article}

\usepackage[]{acl}

\usepackage{times}
\usepackage{latexsym}
\usepackage{comment}
\usepackage{subcaption}
\usepackage{url}
\usepackage{booktabs}
\usepackage{ulem}
\usepackage{amsmath}
\usepackage{latexsym}
\usepackage{multirow}
\usepackage{tabularx}
\usepackage{supertabular}
\usepackage{subcaption}
\usepackage{arydshln} 

\newcommand\tHeadfont{\bfseries}

\usepackage{booktabs} 
\usepackage{makecell} 
\usepackage{siunitx} 

\usepackage{booktabs,xcolor,siunitx}
\usepackage{colortbl} 
\definecolor{lightgray}{gray}{0.9}
\usepackage[T1]{fontenc}
\usepackage[utf8]{inputenc}

\usepackage{microtype}

\usepackage{inconsolata}

\usepackage{multirow}
\usepackage{graphicx}

\def\originalDS{Termite}

\title{Investigating the Impact of Data Contamination of Large Language Models in Text-to-SQL Translation}

\author{
\textbf{Federico Ranaldi$^1$, 
Elena Sofia Ruzzetti$^1$, 
Dario Onorati$^2$}\\
\textbf{Leonardo Ranaldi$^{3,1}$, Cristina Giannone$^4$, Andrea Favalli$^4$}\\
\textbf{Raniero Romagnoli$^4$, Fabio Massimo Zanzotto$^1$} \\
$^1$University of Rome Tor Vergata, Italy \\
$^2$University of Rome La Sapienza, Italy \quad
$^3$Idiap Research Institute, Switzerland \\
$^4$Almawave S.p.A., Via di Casal Boccone, 188-190 00137, Rome, IT
\\
\small{
\href{mailto:federico.ranaldi@alumni.uniroma2.eu}{\color{black} \tt federico.ranaldi@alumni.uniroma2.eu}}
\\
\small{\href{mailto:fabio.massimo.zanzotto@uniroma2.it}{\color{black} \tt fabio.massimo.zanzotto@uniroma2.it}}
}

\begin{document}
\maketitle
\begin{abstract}

Understanding textual description to generate code seems to be an achieved capability of instruction-following Large Language Models (LLMs) in zero-shot scenario. However, there is a severe possibility that this translation ability may be influenced by having seen target textual descriptions and the related code. This effect is known as \textit{Data Contamination}. 

In this study, we investigate the impact of Data Contamination on the performance of GPT-3.5 in the Text-to-SQL code-generating tasks. Hence, we introduce a novel method to detect Data Contamination in GPTs and examine GPT-3.5's Text-to-SQL performances using the known Spider Dataset and our new unfamiliar dataset Termite. Furthermore, we analyze GPT-3.5's efficacy on databases with modified information via an adversarial table disconnection (ATD) approach, complicating Text-to-SQL tasks by removing structural pieces of information from the database. Our results indicate a significant performance drop in GPT-3.5 on the unfamiliar Termite dataset, even with ATD modifications, highlighting the effect of Data Contamination on LLMs in Text-to-SQL translation tasks.

\end{abstract}

\section{Introduction} %->->LEO,FEDE
%LLMs as Code Writers
%Large Language Models (LLMs) have demonstrated remarkable capabilities as semantic interpreters of textual description to generate code for a variety of programming languages 
Large Language Models (LLMs) have been demonstrating to be able to understand the semantics of text descriptions for generating code for a variety of programming languages
\cite{wang2023codet5,zhang-etal-2023-self,chen2023program}. This capability showcases an impressive understanding of the syntax and semantics of both natural language and programming languages. Beyond capturing and mastering the grammar of programming languages governed by a finite set of rules, these models are also proficient in semantically interpreting natural language descriptions and then translating them into a code snippet \cite{yuan2023evaluating}.

%LLMs as SQL Code Writers: Additional Challenges  
LLMs are successful code generators even for the challenging generation of SQL queries from textual description \cite{rajkumar2022evaluating,gao2023texttosql,pourreza2023dinsql}. Indeed, query languages like SQL pose additional challenges as code snippets depend on underlying databases.
%Languages such as SQL are used primarily for querying and managing databases
Then, it is crucial to thoroughly understand the specific database structure with which the SQL code will interact. This is because the effectiveness and accuracy of the generated SQL code largely depend on how well it aligns with the database's schema, constraints, and data types.

%LLMs have shown high proficiency in generating translations from natural language queries to SQL queries that are, in most cases, syntactically correct \cite{rajkumar2022evaluating}. Even when translations are not entirely accurate in terms of fulfilling the user's intended query, queries tend to be at least executable within an SQL environment. Indeed, most errors committed by models in Text-to-SQL translation are of a semantic nature. These results reveal the crucial importance of the information provided about the database. 

%Instillare il dubbio: e se fosse tutto derivante dalla DATA CONTAMINATION
However, the evaluation of the LLMs' capability to generate SQL may be conflated by \textit{data contamination} \cite{magar2022data,ranaldi-etal-2023-precog}.
Data Contamination refers to the situation where a model may have been exposed to, or trained on, parts of the dataset that are later used for its evaluation.
Indeed, many datasets that are used to evaluate LLMs' ability to generate SQL code from text, like Spider \cite{yu2019spider}, may be included in the pre-training material of state-of-the-art instruction following LLMs such as OpenAI GPTs \cite{OpenAI-GPTs}.   
% Data Contamination \cite{magar2022data} refers to the situation where a model may have been exposed to, or trained on, parts of the dataset that are later used for its evaluation. In particular, \citet{magar2022data} and \cite{ranaldi-etal-2023-precog} have tried to detect Data Contamination in BERT \cite{devlin-etal-2019-bert} whose training set is quite known, with targeted techniques and have shown that this problem exists.

In this paper, we aim to unravel the complicated question of whether memorization is responsible for text-to-SQL code generation capabilities of LLMs. More specifically, we focus on the following research questions:
\begin{itemize}
    \item \textit{RQ1}: Is it possible to determine data contamination by solely analyzing the inputs and outputs of existing LLMs?
    \item \textit{RQ2}: Do recent GPTs excel in Text-to-SQL tasks in a zero-shot setting both on potentially leaked data and totally unseen one? %tra dati noti e dati non noti
    \item \textit{RQ3}: Is data contamination affecting the accuracy and reliability of an existing GPT in Text-to-SQL tasks?
\end{itemize} 
%Hence, we propose: (1) \originalDS{} - a fresh dataset for evaluating the task of text-to-SQL - in contrast with the widely spread Spider \cite{yu2019spider}, which has been possibly used for pretraining LLMs such as the commercial GPTs; (2) a measure to determine data contamination in LLMS for tasks of text-to-SQL.
Hence, we propose \originalDS{} - a fresh dataset for evaluating the task of text-to-SQL - in contrast to the widely spread Spider \cite{yu2019spider}, which has possibly been used to pre-train LLMs such as commercial GPTs.
% [RQ1]
By comparing \originalDS{} and Spider, we propose a measure to determine data contamination in LLMs for tasks of text-to-SQL  (\textit{RQ1}).
% [RQ2]
We then experimented with GPT-3.5 \cite{OpenAI-GPT-3.5-turbo}, comparing the results in the text-to-SQL task obtained on \originalDS{} and on Spider (\textit{RQ2}).
% We analyzed the data contamination on the two datasets and removed structural information from tables to obtain more evident results on Spider. 
% [RQ3]
Then, we tested GPT-3.5 by removing structural information from the databases and demonstrate that the model is more resistant to this adversarial input perturbation on leaked data than unseen one (\textit{RQ3}). 
%Results show that not only does GPT exhibit clear knowledge about Spider, but this also leads to an overestimation of the model's performance in Text-to-SQL tasks in zero-shot scenarios.

Our results show that not only does GPT exhibit clear knowledge about Spider, but this also leads to an overestimation of the model's performance in Text-to-SQL tasks in zero-shot scenarios.

\section{Background} %->->LEO,PROF,FEDE,ELENA
\label{sec:background}

Text-to-SQL represents a cutting-edge task in Natural Language Processing (NLP), where the goal is to translate user queries expressed in natural language into SQL queries that can be executed on a database. This task is crucial in making database interactions more accessible to users who may not be familiar with SQL syntax.

In the early stages of Text-to-SQL research, the focus was primarily on rule-based and heuristic approaches \cite{warren-pereira-1982-efficient,giordani-moschitti-2012-translating}. 
%These initial methods relied on manually crafted rules and templates to interpret and translate natural language into SQL. Although effective to some extent, these approaches were limited in their ability to handle the complexity and variability of natural language.

The landscape of Text-to-SQL began to evolve significantly with the advent of neural network-based approaches \cite{yin-etal-2016-neural,xu2017sqlnet}. The shift towards neural models was facilitated by the creation and availability of large, specialized datasets such as Spider \cite{yu2019spider}, which provided diverse and complex natural language to SQL examples. 

The most recent advancements in Text-to-SQL involve the use of Large Language Models (LLMs), which have demonstrated remarkable capabilities in handling various tasks without the need for specific pretraining or fine-tuning tailored to each task.
%Most studies, like \citet{gao2023texttosql} and \citet{pourreza2023dinsql}, focused on evaluating LLMs in Text-to-SQL, apply the Spider Dataset, widely acknowledged as an effective benchmark for assessing performance in this specific task. 
\citet{gao2023texttosql} and \citet{pourreza2023dinsql} have shown that GPTs are effective Text-to-SQL coders on Spider, widely acknowledged as an effective benchmark for assessing performance in this specific task.
On the same dataset, approaches that involve deconstructing the problem in smaller ones via in-context learning \cite{pourreza2023dinsql, zhang2023actsql} are also effectively explored.
However, while LLMs performances have been explored in detail, it remains unclear whether the results may be conflated by data contamination.
Indeed, if it turns out that LLMs perform better on tasks with data that have already been seen during the pretraining phase, we would be facing an issue of data contamination. 

% Since we aim to analyze all these aspects by analyzing Text-to-SQL's performances of GPT-3.5, for which there is no available accurate information about its pretraining data, we require a method to determine whether certain data are known or unknown to the model.

Data Contamination is a relatively new and tricky problem in the field of machine learning, and there are only a few studies that have addressed it. \citet{magar2022data} attempts to examine how accuracies achieved by BERT \cite{devlin-etal-2019-bert} on certain tasks vary from previously seen data e and unseen when the training set contains a portion of the test set.
Recently, the effect of data contamination on BERT and GPT-2 performance on NLU datasets has been discussed by training a model from scratch and measuring the difference in performance over seen and unseen data \cite{ranaldi-etal-2023-precog,jiang2024investigating}.
This line of research is complementary to the one we are proposing in this paper.
In fact, experimenting with very large language models is still challenging.
%-- when not simply unfeasible-- for many academic environments.
These technical limitations lead to experiments that involve training on smaller networks, which resemble the original one but are trained on fewer data and have fewer parameters (as done both in \citet{ranaldi-etal-2023-precog} and \citet{jiang2024investigating}).
Hence, a different strategy is needed to address data contamination in closed models. Like \citet{carlini2021extracting}, we are trying to extract pretraining data information from LLMs, while no accurate information on pretraining data is available.
The concern about Data Contamination is growing along with the popularity of closed LLMs \cite{sainz2023nlp} and some efforts -- like the Contamination Index\footnote{\href{https://hitz-zentroa.github.io/lm-contamination/}{https://hitz-zentroa.github.io/lm-contamination/}} -- are made to trace back training data.

Our work contributes to understanding how data contamination -- also called "Memorization" by \citet{magar2022data}  -- plays a role in Text-to-SQL tasks on black-box models, without any further training step. 
%More precisely, we want to understand whether GPT-3.5 is familiar with Spider Validation Set that will be used further for evaluating Text-to-SQL performances of GPT-3.5. 
In particular, we will test GPT-3.5 on a well-known dataset --Spider-- and compare the performance it achieves on this dataset to that obtained on a new, totally unseen one.
%Thus, taking inspiration from very recent work dealing with Data Contamination in GPT-3.5 \cite{golchin2023time,chang2023speak,deng2023investigating}, we will design specific, tailored tasks on which data contamination and its effect on model performance will be tested.
Thus, taking inspiration from very recent work dealing with Data Contamination in GPT-3.5 \cite{golchin2023time,chang2023speak,deng2023investigating}, we will design specific tasks to assess the presence of data contamination and its effect on model performance.

%
%
% SI CITANO UN PO' DI COSE GIÀ CITATE NEL PAPER DI CLIC (COSE CHE RIGURDANO SPIDER E LA TEXT-TO-SQL)
% SI CITANO COSE CHE RIGUARDANO I NUOVI ASPETTI QUINDI DATA CONTAMINATION E DATABASE PERTURBATION
% DATA CONTAMINATION PAPER 1 ->  https://arxiv.org/pdf/2308.08493.pdf
% DATA CONTAMINATION PAPER 2 -> https://aclanthology.org/2022.acl-short.18.pdf
% PRECOG PAPER ( Leo , Elena , Prof)
% INDICE DELLE CONTAMINATION EVIDENCES -> https://hitz-zentroa.github.io/lm-contamination/
% DATABASE PERTURBATION PAPER 1 ->  https://aclanthology.org/2022.acl-long.142.pdf
%
%
\section{Text-to-SQL Datasets}
\label{sec:dataset}
%To explore whether recent GPTs excel in Text-to-SQL tasks [RQ1] and whether data contamination is responsible for this performance [RQ2], the first step is to introduce the used datasets.
To explore whether some test dataset has been leaked during training (\textit{RQ1}), meausure GPTs performance in Text-to-SQL tasks both on potentially known and unknown data (\textit{RQ2}) and whether data contamination is responsible for this performance (\textit{RQ3}), the first step is to introduce the used datasets.
%The first step is to introduce the used datasets, in order to explore whether recent GPTs excel in Text-to-SQL tasks [RQ1] and whether data contamination is responsible for this performance [RQ2].
In addition to the de-facto standard of Spider \cite{yu2019spider} (described in Sec. \ref{sec:spider}), we propose \originalDS{}, a Text-to-SQL dataset conceived to be a new and never-seen resource (introduced in Sec. \ref{sec:originalDS}).
Therefore, \originalDS{} lowers the probability of performance boost due to data contamination.

\subsection{Spider: Characteristics and Content}
\label{sec:spider}

Spider \cite{yu-etal-2018-spider} is the de-facto standard for training and testing systems on the Text-to-SQL task. Then, this dataset is used in our study on GPTs and it is used to inspire the construction of our \originalDS{} - Text-to-SQL Repository Made Invisible to Engines.   

Spider appears as a collection of databases and associated sets of pairs of natural language (NL) questions and the corresponding SQL translations. 
Databases are structurally represented inside the dataset in the form of SQL dumps, which include the \texttt{CREATE TABLE}  operations and a limited number of \texttt{INSERT DATA} operations for each table.
%For each of these databases, we have a series of pairs consisting of a natural language (NL) query and its corresponding SQL translation.

%Taking the example of the Spider dataset, our datasets will be composed of instances related to a limited number of databases.

%% queries
%%%% hardness delle query
NL questions are organized into four difficulty levels: \texttt{EASY}, \texttt{MEDIUM}, \texttt{HARD}, and \texttt{EXTRA-HARD}.
%For the definition of the hardness level, we refer to the categorization originally made in Spider \cite{yu-etal-2018-spider}.
The difficulty of an NL question is assessed by considering the corresponding SQL query.% over the specific database.
Hence, the difficulty is correlated with the number and kind of operations that the gold query contains: the presence of \texttt{JOIN} operations, aggregation and \texttt{WHERE} conditions contributes to the hardness of the query.
%: the \texttt{JOIN} operations, aggregations, and \texttt{WHERE} conditions define the hardness of a query in the following way.
\texttt{EASY} queries do not involve more than one table. 
\texttt{MEDIUM} and \texttt{HARD} queries span multiple tables: \texttt{MEDIUM} queries contain only a \texttt{JOIN} or aggregation operation whereas \texttt{HARD} queries  are more complex both in terms of number of \texttt{JOIN} and aggregations.
Finally, \texttt{EXTRA-HARD} queries may contain nested queries, and other operators like \texttt{UNION} and \texttt{INTERSECT} 
%TODO in appendice? sul paper di spider non c'e'
\footnote{More details are available on the \href{https://github.com/taoyds/spider/tree/master/evaluation_examples}{official Spider repository}}.

Since our aim is to evaluate the GPT capabilities in zero-shot scenario, we only considered the validation split of Spider. This portion of the dataset consists of 20 databases and 1,035 pairs of NL-SQL queries distributed on the four difficulty categories (see Tab. \ref{tab:datasets_facts}).
%Its creators categorized the queries to be translated into four level of hardness (EASY,MEDIUM,HARD and EXTRA-HARD) based on the complexity of their SQL translation. These queries are related to databases belonging to different domain categories.

%dire che TERMITE è in anglo-italiano e aggiungere citazione sulle multilingual capabilities di GPT-3.5
\subsection{\originalDS{}: a Text-to-SQL Repository Made Invisible to Engines}
\label{sec:originalDS}
The driving idea for proposing a new dataset for the Text-to-SQL task is to reduce the possibility of boosting performance due to data contamination.
Indeed, publicly available datasets are generally not suitable for this purpose. Novel datasets made available, for example, after training a model that one wishes to test, but which are built from publicly available resources such as Kaggle or Wikipedia (this is the case for recently developed datasets like BIRD \cite{li2023llm} or Spider itself), do not guarantee that they are as new as required. The same issue may also be faced for "hidden" test sets.
Also, since freely available datasets are easily accessed and tracked by engines, if not already contaminated, they are at risk of being contaminated in the near future. 
To address these challenges, we propose \originalDS{}\footnote{The repository will be available \href{https://check-data-in-the-attached-materials}{here} under GPL-3.0 license. To access, use the password "youshallnotpass".}. \originalDS{} aims to be a permanently fresh dataset.
Indeed, our dataset will be invisible to search engines since it is locked under an encryption key that is distributed with the dataset. This trick will reduce the accidental inclusion in a novel training set for commercial or research GPTs.    

%The second dataset was constructed by us, drawing inspiration from the characteristics of Spider. 
Drawing inspiration from the characteristics of Spider, \originalDS{} contains hand-crafted databases in different domains. Each database has a balanced set of NL-SQL query pairs: we defined an average of 5 queries per hardness-level. 
The entire dataset was designed to be comparable to the Spider Validation Set, not only in terms of database characteristics such as size and table count (see Table \ref{tab:datasets_facts}) but also in terms of query difficulty, which was measured using the same definition provided by Spider.
Moreover, as in Spider, during the construction of \originalDS{} we took care to
%make nor ambiguous nor vague 
 write unambiguous, direct NL questions that can be solved by a model relying only on its linguistic proficiency and on an analysis of the schema, with no external knowledge needed.
The style adopted in the NL questions is plain and colloquial
%, as is 
in line with 
the style of Spider's NL questions.
Spider and \originalDS{} are also comparable in terms of number of tables and columns in each dataset.
We curated the column names to make them similar to the ones in Spider, using a similar percentage of abbreviations and compound names (see Table \ref{tab:datasets_facts} and Appendix \ref{app:eq_abbr}).
This equivalence will be crucial to limit the influence of the dataset itself on the following evaluations and will be further explored in Section \ref{sec:dataset_equivalence_test}.

However, there is a significant and fundamental difference between the two datasets, as the \originalDS{} is not openly available on the web or easily retrievable nor built on pre-existing openly available resources.
Therefore, we can be confident that our dataset did not contribute in any way to the pretraining of LLMs. This aspect will be crucial in the next sections, where we will investigate data contamination in GPT-3.5.

\begin{table}[]
    \centering
    \resizebox{0.95\linewidth}{!}{
    \begin{tabular}{p{4cm}|cc}
    %\hline
    \cline{2-3}
               & \multicolumn{2}{c}{Dataset} \\ \cline{2-3}
               & \tHeadfont{Spider}       & \tHeadfont{Termite}      \\ \hline \hline
    \small{\#DB}   & 20           & 10           \\
    \small{avg \#TABLES per DB}      & 4.2          & 4.0          \\
    \small{avg \#COLUMNS per TABLE}     & 5.46         & 5.56         \\
    \small{\#QUERY} & 1035         & 202          \\
    \small{avg \#QUERY per DB}      & 51.75        & 20.2         \\
    \small{avg \#FK/\#COLUMNS per DB}          & 0.16         & 0.13         \\
    \small{avg \#Compound/\#COLUMNS per DB}          & 0.63         & 0.51         \\
    \small{avg \#Abbr/\#COLUMNS per DB}          & 0.10         & 0.12         \\ \hline
    \end{tabular}
    }
    \caption{Spider and \originalDS{} fact sheet. Termite is designed to be comparable to the validation set of Spider. }
    \label{tab:datasets_facts}
\end{table}

%\subsection{Human-judged Hardness Equivalence of \originalDS{} and Spider}
\subsection{Comparing Hardness of \originalDS{} vs. Spider}
\label{sec:dataset_equivalence_test}

An inherently different hardness of \originalDS{} and Spider may cause imbalances during a comparative evaluation of LLMs over different sets. Then, we aimed to produce an \originalDS{} that is as close as possible to Spider.
%We adopted a similarity-by-design approach and, then, we evaluated the similarity in hardness of the produced NL queries between the two datasets.

\originalDS{} is designed to resemble Spider in terms of measurable aspects, like the number of columns and tables per database, as well as the lexicon used in the schema definition.
%or numbers of \texttt{JOIN} operations that are included on average on queries {\color{red} (like discussed in Table \ref{app:tab_detailed_statistics})}.
\begin{comment}
Moreover, as in Spider, during the construction of \originalDS{} we took care to
%make nor ambiguous nor vague 
 write unambiguous, direct 
natural language (NL) queries that can be solved by a model relying only on its linguistic proficiency and on an analysis of the schema, with no external knowledge needed.
Moreover, the style adopted in the NL queries is plain and colloquial
%, as is 
in line with 
the style of Spider's NL queries.   
\end{comment}
However, it remains difficult to quantify via some simple statistics
% how comparable are translations required in natural language, especially when it comes to stating 
how hard it is to understand how to translate a natural language question into an SQL statement.

%For this reason, a human-centered definition of hardness is adopted. 
To compare hardness of \originalDS{} and Spider, we adopted a human-centered definition: 
%The main idea is that 
if humans can translate questions into an SQL queries on both Spider and \originalDS{} with the same level of challenge, then it means that their hardness, at least for a SQL-proficient human annotator, is the same.
Therefore, ten annotators were asked to judge the equivalence in terms of hardness of the SQL translations that compose Spider and \originalDS{} by examining a random sample of queries of both datasets.
To measure the hardness of the two datasets, we designed a simple test. Given a Entity-Relationship schema of a database and a question in natural language, each annotator is asked to choose among three options the correct translation in SQL of the question.
Appendix \ref{app:dataset_equivalence_test} presents details on the construction of the test.

On both Spider and \originalDS{}, taking as join annotation the answer chosen by the majority of annotators leads to almost perfect classification ($0.975$ accuracy on Spider and maximum accuracy on \originalDS{}). The average accuracy per annotator is $0.91 (\pm 0.05)$ on Spider and $0.94 (\pm 0.07)$ on \originalDS{}.
Moreover, Fleiss's Kappa coefficients are rather high ($0.79$ and $0.85$ respectively) for both Spider and \originalDS{}.
Hence, we can conclude that humans do not find a dataset more difficult than the other. Then, the two datasets can be considered equivalent in terms of hardness of translations.

\section{Method: studying Data Contamination and its Effect on the Text-to-SQL Task}
\label{sec:methods}
%% spiegazione del ragionamento
%In this Section, we describe how data contamination issue may play an important role in solve a task like Text-To-Sql. 

Our intuition is that data contamination may play an important role in GPT's performance. 
However, investigating the presence of data contamination in GPT models is extremely difficult if there is no possibility to access training datasets.
Then, data contamination can only be estimated.% by interacting with the GPT. 

To investigate our intuition, we first describe a way to quantify the presence of data contamination on GPT by examining database dumps in the Text-to-SQL datasets (Sec. \ref{sec:dc_method}).
Then, we tested GPT-3.5 on the Text-to-SQL task both on possibly already explored and definitely hidden data (Sec. \ref{sec:t2smethod}). 
We expect a decline in performance when the model is required to make inferences on new data not previously encountered. Finally, we describe an adversarial degradation of the input that makes the task of Text-to-SQL translation harder without prior knowledge (Sec. \ref{sec:atd_method}).
Indeed, we argue that if a model achieves high performance in a task by memorizing previously seen information, reducing the quality of input information would not significantly impact its performance on data it has encountered before.

\subsection{Tracing Data Contamination} %->->FEDE,ELENA
\label{sec:dc_method}
%intro: problema dc e risorse (spider e original)
%exp setup e results data contamination: mlm on dump

%\subsection{Data Contamination in LLMs}
Our aim is to understand whether data contamination may have occurred before testing GPT-3.5 performances on Text-To-SQL task. 
The data contamination issue criticality emerges when a model is inadvertently trained on data that include or overlap with its testing dataset: this issue may lead to skewed performance metrics and a misrepresentation of the model's true capabilities. 
%Despite the importance of the issue, studying in detail the training sets on which LLMs are trained may be too difficult, if not impossible, due to the absence of information -- as in the case of GPT-3.5-- about the sources of training.
%Hence, it is necessary to find indirect measures to assess the presence of data contamination.
For models like GPT-3.5 -- black-box models with scarce information about the sources of training -- it is necessary to find indirect measures to assess the presence of data contamination.

In the specific case of Text-to-SQL, along with the request to translate the query, the models trained on this task are provided with information regarding the database schema. In particular, LLMs may also have been trained on the dumps of databases in the Text-to-SQL datasets. 
% During the translation, a model should rely solely on the one-shot dump provided during prompting. We want to undestand whether is possible  or if it considers something else that it might already have in memory.
Hence, it is possible to assess the presence of data contamination on Text-to-SQL datasets by measuring the previous knowledge that a model has on these dumps.

% precog sul dump
A clue to determine whether the data contamination has occurred is that the model is able to reconstruct missing information regarding the database schema.
Since LLMs are trained to produce text, we propose to measure the accuracy that a model achieves in reconstructing a dump that has been masked.
If the model is able to reconstruct this information on potentially seen data --as Spider's validation dataset might be -- and fails to reconstruct it on the new resources -- like \originalDS{}-- we argue that data contamination has occurred.

In particular, a dump was masked by replacing the 25\% of columns in each table with a [MASK] token. Then, GPT-3.5 was prompted to reconstruct the dump by replacing the masked tokens with appropriate column names.
% The selected prompt is \texttt{``In the following SQL dump there are some [MASK] tokens. Fill with the corresponding column names all the [MASK] tokens. Return the filled dump:''}, followed by the dump of the database with the masked columns. 
In these experiments, the \texttt{INSERT} instructions are also removed from the dump to limit the possible inference regarding the names of columns. 
% We avoid masking the tables primary keys since they are often id and have too regular names.
% va messo?
\begin{comment}
An example of requests is available in Table \ref{tab:es_prompt_dc}.

\begin{table}[h]
\noindent
{\setlength{\fboxsep}{-1.4pt} 
\colorbox{lightgray}{ 
\begin{tabular}[t]{|p{0.45\textwidth}|}
    \hline
    \textbf{user:} In the following SQL dump there are some [MASK] tokens. [...] \texttt{CREATE TABLE "continents" ( "ContId" INTEGER PRIMARY KEY, "[MASK]" TEXT );[...]}  \\
    \hline  
     \texttt{\textbf{GPT-3.5:} CREATE TABLE "continents" ("ContId" INTEGER PRIMARY KEY, "ContinentName" TEXT ); [...]}\\
    
    \hline
    \end{tabular}
    } 
    }
\hfill

\caption{Example of prompt.}
\label{tab:es_prompt_dc}

\end{table}
\end{comment}
%\prompt{In the following SQL dump there are some [MASK] tokens. [...] "CREATE TABLE "continents" ( "ContId" INTEGER PRIMARY KEY, "[MASK]" TEXT );[...]"}{CREATE TABLE "continents" ("ContId" INTEGER PRIMARY KEY, "ContinentName" TEXT ); [...]}
It is important to note that the task is still feasible even if no data contamination has occurred: column names can be deduced from the names of the tables and other columns. However, the task would be much easier in the presence of data contamination.

Hence, given the reconstructed dump from GPT-3.5, we define the DC-accuracy as the percentage of times the predicted column name is equal to the true column name:
\[ \textrm{DC-accuracy} = \frac{\textrm{\# of correct columns name}}{\textrm{\# of columns}}\]
It is possible to assess the presence of data contamination by measuring the DC-accuracy both on the Spider dumps and on the \originalDS{} dump databases.

%%%% Da far salire
\begin{comment}
\subsection{Datasets}
\label{sec:dataset}
In order to compare Text-to-SQL performances on known and unknown datasets, our work has been directed towards finding two datasets for the task that possess certain characteristics. Consequently, we have identified two datasets: Spider and ?Original?.
While the first dataset appears to be known to GPT-3.5, the second, created by us, is not present on the Web and it is impossible for it to have accidentally ended up in the pretraining data of the LLM.
\subsubsection{Spider}
Prompting ChatGPT with generic questions about the dataset, and receiving accurate responses, leads us to consider Spider as a candidate to highlight data contamination for the Text-to-SQL task. Particularly we will study previous knowledge and Text-to-SQL performances exclusively on the Validation Set.
\subsubsection{?Original?}
With the help of human annotators we created a Text-to-SQL dataset containing database whose structural properties are similar to Spider's ones. 
\end{comment}

%
% Entriamo nel merito della Data Contamination
% Spieghiamo effetti della DC nei tasks e quindi in Text-to-SQL
% Cosa possiamo fare in ambito DC rispetto a GPT (limitazioni e proposte)
% Cosa potrebbe ""RICORDARE" GPT 3.5 a proposito di Spider???
% Come rileviamo DC nel nostro caso (MLM sui Dump) ???
% Risultati e considera\newpage %TO BE REMOVEDzioni
%

\subsection{Prompting LLMs for Text-to-SQL Translation}    %->->FEDE , ELENA
\label{sec:t2smethod}
%
% Come affrontiamo il task di T2S
%
% Perche' inviare il dump e chiedere le traduzioni e' ragionevole
Given an instruction in natural language, LLMs can translate the request into code -- and SQL queries, in particular -- to answer the given request.
Specifically, OpenAI's models for generating text
%, commonly referred to as generative pre-trained transformers or large language models, 
have undergone training to process both natural language and code. These models produce text-based outputs as a result of the inputs they receive. 
For this reason, it is possible to frame the Text-to-SQL as a translation task: given a dump for a database and a query in natural language, the model is asked to translate the latter in the corresponding SQL query, referring to tables and columns into the considered database.
The desiderata is an executable query, semantically equivalent to a gold human-generated query.
In the next paragraphs, we first describe how GPT-3.5 -- in particular, \texttt{gpt-3.5-turbo} -- is prompted in order to obtain the translations and then how it is possible to automatically evaluate the performance of this system on both Spider and \originalDS{} datasets.
%Finally, we report the accuracy of GPT-3.5 on Spider and \originalDS{} databases (Section \ref{sec:t2seval}), noticing how, across different query difficulties, the model achieves better performance on seen rather then unseen databases.

%\subsection{Datasets} %->->FEDE,ELENA,DARIO
%
% Spider e Students??? : database e query hardness-level...
%
\paragraph{Text-to-SQL as a Translation Task} %->ELENA, FEDE
%
% GPT-3.5 Turbo 
% API : metodo di interazione
%
OpenAI API's enable to interrogate a model in a multi-turn conversation format: chat models receive a series of messages as input and generate a message as output.
We test the ability of GPT-3.5 on the Text-to-SQL task by framing each translation from natural language to SQL as a separate conversation.

In particular, given a target database, in the first message, the model is given the dump of the database.
In each dump, information about the tables that constitute the database is provided by the \texttt{CREATE TABLE} statements. In the \texttt{CREATE} instructions, the constraints of the primary and foreign keys are also encoded.
In addition, some realistic data to fill the tables are provided by \texttt{INSERT} instructions.
Given the dump, the model answers by producing an interpretation of the dump. Typically, this model response contains an explanation of the contents of the dump.
For example, considering the database \texttt{car\_1} in the Spider dataset, the first messages in the conversation are the following:

\begin{table}[h]
\noindent
{\setlength{\fboxsep}{-1.4pt} 
\colorbox{lightgray}{ 
\begin{tabular}[t]{|p{0.45\textwidth}|}
    \hline
    \textbf{user:} \texttt{CREATE TABLE "continents" [...]; CREATE TABLE "countries" [...];}\\
    \hline  
     \textbf{GPT-3.5:} \texttt{The code above includes the creation of six tables: continents, countries [...]}\\
    
    \hline
    \end{tabular}
    } 
    }
\hfill

%\caption{Example of prompt.}
%\label{tab:es_prompt}

\end{table}

%\prompt{CREATE TABLE "continents" [...]; CREATE TABLE "countries" [...];}{The code above includes the creation of six tables: continents, countries [...]}

Then, given the dump and the interpretation that the model gives of it, a message containing the natural language question to be translated is sent.
In particular, the selected prompt ensures that the model translates natural language questions into SQL queries with a limited amount of text that is not SQL.
%Ci va? ora si legge da esempio
%The selected prompt is: \textit{"Translate in SQL the following query. Answer using only SQL."}
These steps are repeated for each question separately to obtain translations independently of each other.
However, to ensure that the model's understanding of each database is comparable across all questions, the database dump and the same interpretation initially produced by the model are sent as context, in the form of preceding messages, before each translation is requested.
%% alternativa per risparmiare spazio
Hence, building from the previous example, a conversation to translate a question on the \texttt{car\_1} database would be completed by the following messages:

\begin{table}[h]
\noindent
{\setlength{\fboxsep}{-1.4pt} 
\colorbox{lightgray}{ 
\begin{tabular}[t]{|p{0.45\textwidth}|}
    \hline
    \textbf{user:} \texttt{Translate in SQL the
following query. Answer using only SQL. What is the number of continents?}  \\
    \hline
     \textbf{GPT-3.5:} \texttt{SELECT COUNT(*) as n\_conts FROM continents;}\\
    
    \hline
    \end{tabular}
    } 
    }
\hfill

%\caption{Example of prompt.}
%\label{tab:es_prompt}

\end{table}

%\prompt{Translate in SQL the following query. Answer using only SQL. What is the number of continents?}{SELECT COUNT(*) as number\_of\_continents FROM continents;}
%Hence, building from the previous example, a complete conversation to translate a query on the \texttt{car\_1} database is: 
%\prompt{CREATE TABLE "continents" [...]; CREATE TABLE "countries" [...];}{The code above includes the creation of six tables: continents, countries [...]}
%\prompt{Translate in SQL the following query. Answer using only SQL. What is the number of continents?}{SELECT COUNT(*) as number\_of\_continents FROM continents;}
Our approach is completely zero-shot, to minimize the effect that the prompt itself--rather than data contamination--can have on performance.
Once the translation process is completed, the SQL code produced by the model is retrieved to evaluate whether or not the generated query satisfies the natural language query.

\paragraph{Test Suite Accuracy: the Evaluation Metric} %->ELENA,FEDE
%\label{sec:TSA_metric}
%
% Test Suite Accuracy
%
%In this section, we briefly describe the Test Suite Accuracy metric.
We adopted the Test Suite Accuracy metric as an evaluation metric in our experiments.
This score was introduced by \citet{zhong-etal-2020-semantic} and currently is the official metric for the evaluation of systems tested on Spider\footnote{As reported on the \href{https://yale-lily.github.io/spider}{Spider official website}}.
In principle, given a query $q$ generated by a system, one would like to evaluate whether $q$ is semantically equivalent to a human-generated gold query $g$.
This metric, known as semantic accuracy, is undecidable in general \cite{chu2017cosette}.
The Test Suite Accuracy metric aims to approximate semantic accuracy and states the correctness or incorrectness of $q$ by comparing the denotation of a gold query $g$ and the denotation of $q$. However, it introduces fewer false positives than Execution Accuracy -- that compares $g$ and $q$ on a single database -- by comparing the denotation of both queries on as few databases as possible. This set of random databases 
%on which queries are tested 
is called the Test Suite.

In our experiments, a Test Suite of maximum 1000 random databases is constructed for each database upon which queries are defined, as reported in the original paper.
Therefore, a model-generated query $q$ is executed on the Test Suite databases and labeled correct if no database can distinguish it from the gold query $q$.
The 97\% of the queries is automatically evaluated, while the remaining ones are manually evaluated by three SQL-proficient annotators. This step is necessary because, in these rarer cases, the queries -- either gold or model-generated -- make use of functions not available in the SQL server used to execute the queries.

Using the Test Suite Accuracy as an approximation of the semantic accuracy of the system, we show that, among different databases and different query difficulties, GPT-3.5 demonstrates different performance on Spider and \originalDS{} datasets (Section \ref{sec:t2seval}).
In addition, the same type of evaluation will be performed on adversarially degraded databases in Section \ref{sec:atd_results} to establish that the highest performance degradation occurs on unseen data.

\subsection{The Adversarial Table Disconnection} %->ELENA,FEDE,DARIO
\label{sec:atd_method}
%\subsection{Correlation with Data Contamination results} %->ELENA,FEDE,PROF
%
% Proviamo a dare una motivazione si due risultati
% Sono Correlati??? da elena: per il momento no! non c'era correlazione tra risultati per data contamination sui diversi database e gli score ottenuti
% Mette tutto insieme, data la presenza di data contamination, e dopo aver osservato valori di accuracy piu' bassi su database mai visti, allora dimostra che l'effetto e' dato dalla data contamination, mostrando che chat gpt e' piu' resistente a perturbazioni su spider che sugli original
%motivazione
The differences in performance over seen rather than unseen data it is not, in principle, sufficient to state that the observed differences are caused by data contamination issues since it is still possible that the datasets hinder some biases.
On the one hand, we designed the \originalDS{} dataset to be comparable to Spider; hence, once the hardness of the queries is also fixed, performances are comparable.
On the other hand, we want to ensure that memorization is, in fact, playing an important role. For this reason, we propose an adversarial approach to state the importance of memorization on this task. We will refer to this method as Adversarial Table Disconnection (ATD).

%dettagli
The ATD aims to make the translation process from a request in natural language into SQL harder. In particular, ATD disconnects the tables from each other, making it more difficult to figure out on which columns the \texttt{JOIN} needs to be performed.
ATD disconnects tables by removing the foreign keys constraints. In particular, all instructions referring to the creation of the constraint are removed from the dump. This structural information is critical to translating a questions into SQL queries: we argue that the removal of this information has a crucial impact on translation, both for humans and systems, unless the missing information can be retrieved. %% o come farsi menare in tre semplici passi
Given the importance of the presence of a foreign key, our aim is to remove insights about the relation between columns that may be directly used to infer the removed relationship. Hence, ATD involves the removal of all the \texttt{INSERT} instructions from the dump. In fact, matching values into this type of instructions may give direct clues about the columns relationships.
Crucially, our aim is not to change the semantic of the database, but only to make it less easy to understand. For this reason, the original column names are kept, making inferences about the content of the database still possible. 
Hence, after ATD a model is given a dump deprived of structural information, with tables disconnected from one another but still semantically equivalent to the original one.

%risultati attesi
Because ATD makes the task more difficult, a drop in system performance is expected.
However, the drop can be mitigated by relying on prior information about the database structure. Therefore, given the presence of data contamination, we expect GPT-3.5 to be robust to ATD perturbation on Spider datasets, with a more pronounced performance loss on \originalDS{} databases.

\section{Experiments}
\label{sec:experiments}

%%%%% Metriche ALT
% accuracy totali
% spider su 138 predizioni 0.35507246376811596
% original-hard-reduced su 66 0.15151515151515152

\subsection{Quantifying the Data Contamination in Text-to-SQL datasets}
\label{sec:dc_results}
%Data Contamination results
It is possible to quantify the presence of data contamination by comparing the DC-accuracy that GPT-3.5 achieves in predicting column names on a new dumps --from \originalDS{}-- versus potentially already seen dumps in Spider (as described in Section \ref{sec:dc_method}).
In Table \ref{tab:dc_results}, the average DC-accuracy over Spider and \originalDS{} datasets is reported. 

The model seems to find the task easier on Spider databases than on \originalDS{} ones.
In particular, the average accuracy of Spider dumps is more than $33\%$, that is, on average, more than $20\%$ higher than the score on \originalDS{}.
Moreover -- while on both datasets, some databases are hard to predict, with a minimum accuracy of $0$  -- on the Spider dataset, GPT-3.5 achieves a perfect accuracy on two databases. The same does not hold for \originalDS{}, where the highest accuracy is $44\%$.
The different performance of the model on these two datasets suggests the presence of data contamination.

\begin{table}[]
\centering
    \begin{tabular}{
      l
      l
      l
    }
    \toprule
    DC-accuracy      & \tHeadfont{Spider} & \tHeadfont{\originalDS{}} \\
    \midrule
    Mean  & $33.42(\pm 33.01)$  & $13.21(\pm 18.70)$ \\
    Min - Max   & $0.00 - 100.00$  & $0.00 - 44.44$  \\
    \bottomrule
    \end{tabular}
    \caption{Average, min, and max accuracies of GPT-3.5 on predicting the masked columns names on dumps in Spider and \originalDS{}. The overall performances in terms of DC-accuracy over the Spider dataset are superior with respect to the one that can be observed on \originalDS{} dataset.}
    \label{tab:dc_results}
\end{table}
% va detto quali?
%In particular, GPT-3.5 achieves a perfect score on two databases of the Spider dataset.
It is also interesting to notice that on 35\% dumps (7 dumps), the DC-accuracy is over 40\%, while only on two databases among the ones in \originalDS{} GPT-3.5 achieves (with a score of 44.44\% and 40\%) the same results. 
%% TODO manca qualcosa qui? si fa in tempo a fare analisi su quanti sono nomi composti ad esempio?
A complete list of accuracies per database can be found in Appendix \ref{app:dc_db_results}
The different performance in terms of DC-accuracy over \originalDS{} with respect to Spider suggests the presence of data contamination.

\begin{table*}[]
\centering
\begin{tabular}{
  l
  S[table-format=2.2(4)]
  S[table-format=2.2(4)]
  S[table-format=2.2(4)]
  S[table-format=2.2(4)]
}
\toprule
\multirow{2}{*}{\tHeadfont{Hardness}} & \multicolumn{2}{c}{Original Dumps} & \multicolumn{2}{c}{Adversarial Table Disconnection} \\
\cmidrule(lr){2-3} \cmidrule(lr){4-5}
& {\tHeadfont{Spider}} & {\tHeadfont{\originalDS{}}} & {\tHeadfont{Spider}} & {\tHeadfont{\originalDS{}}} \\
\midrule
\texttt{EASY} &  {$90.11(\pm 11.65)$} & {$74.00(\pm 21.19)$} & {$91.08(\pm 10.32)$} & {$62.00(\pm 19.89)$} \\
\texttt{MEDIUM} & {$77.21(\pm 16.35)$} & {$67.06(\pm 24.83)$} & {$72.71(\pm 23.63)$} & {$63.70(\pm 16.03)$}  \\
\texttt{HARD}    & {$48.83(\pm 23.17)$} & {$28.33(\pm 23.78)$} & {$48.71(\pm 28.79)$} & {$22.67(\pm 22.20)$}  \\
\texttt{EXTRA-HARD}   & {$30.94(\pm 23.79)$} & {$31.14(\pm 24.54)$} & {$28.96(\pm 19.28)$} & {$28.98(\pm 17.03)$} \\
\bottomrule
\end{tabular}
\caption{GPT-3.5 accuracy on the Spider Dataset and the \originalDS{} Dataset, across four levels of hardness of queries. The results reported are average accuracy across all databases in the two datasets. The first two columns refer to accuracies on the standard task, while the last columns show results after ATD.}
\label{tab:t2sacc}
\end{table*}

\begin{comment}
\begin{table*}[]
\begin{tabular}{|l|ll|ll|}
\hline
\multirow{2}{*}{hardness} & \multicolumn{2}{l|}{Unalterd Dumps} & \multicolumn{2}{l|}{Adversarial Table Disconnection} \\ \cline{2-5} 
                          & \multicolumn{1}{l|}{Spider} & our   & \multicolumn{1}{l|}{Spider}          & our           \\ \hline
easy                      & \multicolumn{1}{l|}{87.10}  & 71.74 & \multicolumn{1}{l|}{87.10}           & 58.70         \\ \hline
medium                    & \multicolumn{1}{l|}{75.56}  & 66.04 & \multicolumn{1}{l|}{71.97}           & 62.26         \\ \hline
hard                      & \multicolumn{1}{l|}{51.72}  & 29.73 & \multicolumn{1}{l|}{50.00}           & 24.32         \\ \hline
extra                     & \multicolumn{1}{l|}{29.52}  & 31.58 & \multicolumn{1}{l|}{25.30}           & 29.82         \\ \hline
\end{tabular}
\end{table*}
\end{comment}

\subsection{Measuring GPT-3.5 performances on seen and unseen data}
\label{sec:t2seval}
% risultati prima di ATP
% Experiments
% Analisi parallela sui due Datasets per hardness level
% Confronto tra le avg accuracy dei due datasets
Having estimated the presence of data contamination, we focus on the analysis of the performance of GPT-3.5 on the dataset presented in Section \ref{sec:dataset}.
The results described here suggest the role that memorization may play in the performance of a Large Language Model like GPT-3.5.
We analyze the model's performance by categorizing queries according to their hardness and averaging across the different databases of the two datasets (see Appendix \ref{app:t2seval_dbdetail} for the results on different databases).

\begin{comment}
\begin{table}[h!]
\begin{center}
\begin{tabular}{|l|l|l|}
\hline
Hardness & Spider & \originalDS{} \\ \hline 
\texttt{EASY}       & $90.11(\pm 11.65)$ & $74.00(\pm 21.19)$  \\ \hline
\texttt{MEDIUM}     & $77.21(\pm 16.35)$ & $67.06(\pm 24.83)$ \\ \hline
\texttt{HARD}       & $48.83(\pm 23.17)$ & $28.33(\pm 23.78)$ \\ \hline
\texttt{EXTRA-HARD}      & $30.94(\pm 23.79)$ & $31.14(\pm 24.54)$ \\ \hline
\end{tabular}
\end{center}
\caption{GPT-3.5 accuracy on the Spider Dataset on the \originalDS{} Dataset, across four levels of hardeness of queries. The results here reported are average accuracy across all the databases in the two dataset.}
\label{tab:t2sacc}
\end{table}
\end{comment}

Table \ref{tab:t2sacc} reports the average Test Suite Accuracy results for each hardness level. 
We notice that, on both sets of databases, the accuracy of the model decreases as the hardness increases. 
In particular, the \texttt{EASY} queries on the Spider dataset achieves, on average, accuracy over the $90\%$. 
Accuracy decreases progressively, with the greatest drop ($29\%$) between \texttt{MEDIUM} and \texttt{HARD} levels.
The worst accuracy is obtained on the \texttt{EXTRA-HARD} queries.
The same trend is also observed on the \originalDS{} dataset: on the queries \texttt{EASY} GPT-3.5 achieves an average accuracy of $74\%$. Again, a decrease in performance is observed on the \texttt{MEDIUM} and \texttt{HARD} queries, while on \originalDS{} \texttt{EXTRA-HARD} queries GPT-3.5 appears to achieve performance similar to \texttt{HARD} queries.

However, comparing the results on the two datasets, it is possible to notice that, given a certain hardness level, the accuracy of GPT-3.5 is not comparable on the two datasets.
%% facciamo stat significance? sign t??
In fact, the average performance difference between Spider and the \originalDS{} dataset is remarkable:
\texttt{EASY} query accuracy decreases by $16\%$, $10\%$ for \texttt{MEDIUM} ones. The biggest drop is observed on \texttt{HARD} queries, with a $20\%$ difference in performance.
The only accuracies that appear to be similar -- and sensibly lower -- are on \texttt{EXTRA-HARD} hard queries.

These results provide insight that GPT-3.5 capability on the task may be highly influenced by data contamination issues. In fact, given comparable queries from a hardness perspective, results on databases that have never been seen turn out to be worse than those that have already been made available and, likely, observed in training.

\subsection{Robustness of GPT-3.5 on Text-to-SQL performances after ATD on seen data} %->ELENA,FEDE
\label{sec:atd_results}
%
% Vediamo risultati Text-to-SQL su db perturbati e li compariamo con quelli precedenti
%
To better understand whether the data contamination is responsible for the difference in performance observed in Table \ref{tab:t2sacc}, we analyze the accuracy over Spider and \originalDS{} after ATD.

As expected, a greater performance drop is observed over \originalDS{} databases, while the model seems to be robust against the ATD over the Spider dataset.
In particular, the accuracy over the \texttt{EASY} queries decreases by $12$ points on average on the \originalDS{} dataset, while similar results (close to the $90\%$) can be observed in Spider.
On the \texttt{MEDIUM} queries, a slightly more pronounced difference in performances can be observed over Spider ($4.5$ points) with respect to the one observed in \originalDS{} ($3.36$ points). 
It is on the \texttt{HARD} queries, however, that the different performances on seen and unseen data are much more evident. Those queries require more \texttt{JOIN} operations than the previous ones. On the one hand, on the Spider databases, the average performance is around $48\%$ for both the original dumps and the dumps on which ATD is applied. 
On the other hand, an average performance drop of $5.66$ points is observed on the \originalDS{} dumps. 
Finally, similar and generally lower performances can be observed over the \texttt{HARD} queries.

% molto dura
Hence, this final experiment confirms that --since the drop observed in the performance of GPT-3.5 after ATD is greater on new data than on contaminated ones -- the memorization ability of the model plays a crucial role in its performance.

\begin{comment}
\begin{table}[]
\begin{tabular}{|l|c|c|}
\hline
hardness                             & Spider                          & \originalDS{}                             \\ \hline
\multirow{2}{*}{easy ATD}   & 91.08(\pm 10.32) & 62.0(\pm 19.89)  \\ 
                                     & + 0.97  & - 7.00   \\ \hline
%                                     & 90.11(\pm 11.65)  & 74.0(\pm 21.19)  \\ \hline
\multirow{2}{*}{medium ATD} & 72.71(\pm 23.63) & 63.7(\pm 16.03)  \\  
                                     & 77.21(\pm 16.35) & 67.06(\pm 24.83) \\ \hline
\multirow{2}{*}{hard ATD}   & 48.71(\pm 28.79) & 22.67(\pm 22.2)  \\  
                                     & 48.83(\pm 23.17) & 28.33(\pm 23.78) \\ \hline
\multirow{2}{*}{extra ATD}  & 28.96(\pm 19.28) & 28.98(\pm 17.03) \\ 
                                     & 30.94(\pm 23.79) & 31.14(\pm 24.54) \\ \hline
\end{tabular}
\end{table}
\end{comment}

\section{Conclusions}

This paper shows that data contamination is responsible for overestimating the performance of GPT-3.5 on Text-to-SQL. The experiments conducted, using a novel metric for detecting data contamination, clearly demonstrate that GPT-3.5 possesses prior knowledge on the contents of the Spider validation set in contrast to his ignorance of our constructed Text-to-SQL unseen dataset, Termite.
In fact, as results show, Text-to-SQL performances on Spider are significantly better than on Termite. This suggests that GPT-3.5 capabilities in zero-shot scenario might not be as surprising as previously thought. Observing the results of data contamination alongside with performances achieved in Text-to-SQL on the two datasets, we concluded that it is indeed the prior knowledge of GPT-3.5 on the test set that makes a significant difference. In addition to this, we found that Adversarial Table Disconnection impacts the results of Text-to-SQL tasks differently across datasets: its influence is relatively mild in the case of the Spider dataset but more pronounced with the Termite dataset. 

Since data contamination is the main responsible for overestimating performances on Text-to-SQL and, possibly, on other tasks, a more thorough reexamination of current LLM's benchmarks for downstream tasks in zero-shot scenarios would be needed. Furthermore, it would be beneficial to develop public datasets, like our Termite, that remain outside the LLM's pretraining. This may guarantee that evaluations on pretrained LLMs are not impacted by Data Contamination.

\section*{Limitations}

Our analysis of data contamination of GPTs has some limitations. Below, we describe some of these and suggest directions for
future work

First, the impact of Data Contamination on the performance of Text-to-SQL tasks has been tested specifically on GPT-3.5. This is a limitation and the analysis should be extended to other models. However, we performed preliminary small-scale pilot experiments akin to those conducted in this study. Results suggest that Data Contamination also affects GPT-4. 

Furthermore, we used only a public dataset for this task. However, this single dataset already shows that data contamination is a relevant issue in measuring performance.  

\section*{Acknowledgements}

% Entries for the entire AnthologyreallloIn fact ,surprising  shows ,ted bries
\bibliography{anthology,custom}

\begin{thebibliography}{28}
\expandafter\ifx\csname natexlab\endcsname\relax\def\natexlab#1{#1}\fi

\bibitem[{Carlini et~al.(2021)Carlini, Tramer, Wallace, Jagielski, Herbert-Voss, Lee, Roberts, Brown, Song, Erlingsson, Oprea, and Raffel}]{carlini2021extracting}
Nicholas Carlini, Florian Tramer, Eric Wallace, Matthew Jagielski, Ariel Herbert-Voss, Katherine Lee, Adam Roberts, Tom Brown, Dawn Song, Ulfar Erlingsson, Alina Oprea, and Colin Raffel. 2021.
\newblock \href {http://arxiv.org/abs/2012.07805} {Extracting training data from large language models}.

\bibitem[{Chang et~al.(2023)Chang, Cramer, Soni, and Bamman}]{chang2023speak}
Kent~K. Chang, Mackenzie Cramer, Sandeep Soni, and David Bamman. 2023.
\newblock \href {http://arxiv.org/abs/2305.00118} {Speak, memory: An archaeology of books known to chatgpt/gpt-4}.

\bibitem[{Chen et~al.(2023)Chen, Ma, Wang, and Cohen}]{chen2023program}
Wenhu Chen, Xueguang Ma, Xinyi Wang, and William~W. Cohen. 2023.
\newblock \href {http://arxiv.org/abs/2211.12588} {Program of thoughts prompting: Disentangling computation from reasoning for numerical reasoning tasks}.

\bibitem[{Chu et~al.(2017)Chu, Wang, Weitz, and Cheung}]{chu2017cosette}
Shumo Chu, Chenglong Wang, Konstantin Weitz, and Alvin Cheung. 2017.
\newblock Cosette: An automated prover for sql.
\newblock In \emph{CIDR}.

\bibitem[{Deng et~al.(2023)Deng, Zhao, Tang, Gerstein, and Cohan}]{deng2023investigating}
Chunyuan Deng, Yilun Zhao, Xiangru Tang, Mark Gerstein, and Arman Cohan. 2023.
\newblock \href {http://arxiv.org/abs/2311.09783} {Investigating data contamination in modern benchmarks for large language models}.

\bibitem[{Devlin et~al.(2019)Devlin, Chang, Lee, and Toutanova}]{devlin-etal-2019-bert}
Jacob Devlin, Ming-Wei Chang, Kenton Lee, and Kristina Toutanova. 2019.
\newblock \href {https://doi.org/10.18653/v1/N19-1423} {{BERT}: Pre-training of deep bidirectional transformers for language understanding}.
\newblock In \emph{Proceedings of the 2019 Conference of the North {A}merican Chapter of the Association for Computational Linguistics: Human Language Technologies, Volume 1 (Long and Short Papers)}, pages 4171--4186, Minneapolis, Minnesota. Association for Computational Linguistics.

\bibitem[{Gao et~al.(2023)Gao, Wang, Li, Sun, Qian, Ding, and Zhou}]{gao2023texttosql}
Dawei Gao, Haibin Wang, Yaliang Li, Xiuyu Sun, Yichen Qian, Bolin Ding, and Jingren Zhou. 2023.
\newblock \href {http://arxiv.org/abs/2308.15363} {Text-to-sql empowered by large language models: A benchmark evaluation}.

\bibitem[{Giordani and Moschitti(2012)}]{giordani-moschitti-2012-translating}
Alessandra Giordani and Alessandro Moschitti. 2012.
\newblock \href {https://aclanthology.org/C12-2040} {Translating questions to {SQL} queries with generative parsers discriminatively reranked}.
\newblock In \emph{Proceedings of {COLING} 2012: Posters}, pages 401--410, Mumbai, India. The COLING 2012 Organizing Committee.

\bibitem[{Golchin and Surdeanu(2023)}]{golchin2023time}
Shahriar Golchin and Mihai Surdeanu. 2023.
\newblock \href {http://arxiv.org/abs/2308.08493} {Time travel in llms: Tracing data contamination in large language models}.

\bibitem[{Jiang et~al.(2024)Jiang, Liu, Zhong, Schaeffer, Ouyang, Han, and Koyejo}]{jiang2024investigating}
Minhao Jiang, Ken~Ziyu Liu, Ming Zhong, Rylan Schaeffer, Siru Ouyang, Jiawei Han, and Sanmi Koyejo. 2024.
\newblock \href {http://arxiv.org/abs/2401.06059} {Investigating data contamination for pre-training language models}.

\bibitem[{Li et~al.(2023)Li, Hui, Qu, Yang, Li, Li, Wang, Qin, Cao, Geng, Huo, Zhou, Ma, Li, Chang, Huang, Cheng, and Li}]{li2023llm}
Jinyang Li, Binyuan Hui, Ge~Qu, Jiaxi Yang, Binhua Li, Bowen Li, Bailin Wang, Bowen Qin, Rongyu Cao, Ruiying Geng, Nan Huo, Xuanhe Zhou, Chenhao Ma, Guoliang Li, Kevin C.~C. Chang, Fei Huang, Reynold Cheng, and Yongbin Li. 2023.
\newblock \href {http://arxiv.org/abs/2305.03111} {Can llm already serve as a database interface? a big bench for large-scale database grounded text-to-sqls}.

\bibitem[{Magar and Schwartz(2022)}]{magar2022data}
Inbal Magar and Roy Schwartz. 2022.
\newblock \href {http://arxiv.org/abs/2203.08242} {Data contamination: From memorization to exploitation}.

\bibitem[{OpenAI(2023{\natexlab{a}})}]{OpenAI-GPT-3.5-turbo}
OpenAI. 2023{\natexlab{a}}.
\newblock \href {https://platform.openai.com/docs/models/gpt-3-5} {Gpt-3.5turbo}.

\bibitem[{OpenAI(2023{\natexlab{b}})}]{OpenAI-GPTs}
OpenAI. 2023{\natexlab{b}}.
\newblock \href {https://platform.openai.com/docs/models} {Gpt's family}.

\bibitem[{Pourreza and Rafiei(2023)}]{pourreza2023dinsql}
Mohammadreza Pourreza and Davood Rafiei. 2023.
\newblock \href {http://arxiv.org/abs/2304.11015} {Din-sql: Decomposed in-context learning of text-to-sql with self-correction}.

\bibitem[{Rajkumar et~al.(2022)Rajkumar, Li, and Bahdanau}]{rajkumar2022evaluating}
Nitarshan Rajkumar, Raymond Li, and Dzmitry Bahdanau. 2022.
\newblock \href {http://arxiv.org/abs/2204.00498} {Evaluating the text-to-sql capabilities of large language models}.

\bibitem[{Ranaldi et~al.(2023)Ranaldi, Ruzzetti, and Zanzotto}]{ranaldi-etal-2023-precog}
Leonardo Ranaldi, Elena~Sofia Ruzzetti, and Fabio~Massimo Zanzotto. 2023.
\newblock \href {https://aclanthology.org/2023.ranlp-1.103} {{P}re{C}og: Exploring the relation between memorization and performance in pre-trained language models}.
\newblock In \emph{Proceedings of the 14th International Conference on Recent Advances in Natural Language Processing}, pages 961--967, Varna, Bulgaria. INCOMA Ltd., Shoumen, Bulgaria.

\bibitem[{Sainz et~al.(2023)Sainz, Campos, García-Ferrero, Etxaniz, de~Lacalle, and Agirre}]{sainz2023nlp}
Oscar Sainz, Jon~Ander Campos, Iker García-Ferrero, Julen Etxaniz, Oier~Lopez de~Lacalle, and Eneko Agirre. 2023.
\newblock \href {http://arxiv.org/abs/2310.18018} {Nlp evaluation in trouble: On the need to measure llm data contamination for each benchmark}.

\bibitem[{Wang et~al.(2023)Wang, Le, Gotmare, Bui, Li, and Hoi}]{wang2023codet5}
Yue Wang, Hung Le, Akhilesh~Deepak Gotmare, Nghi D.~Q. Bui, Junnan Li, and Steven C.~H. Hoi. 2023.
\newblock \href {http://arxiv.org/abs/2305.07922} {Codet5+: Open code large language models for code understanding and generation}.

\bibitem[{Warren and Pereira(1982)}]{warren-pereira-1982-efficient}
David~H.D. Warren and Fernando~C.N. Pereira. 1982.
\newblock \href {https://aclanthology.org/J82-3002} {An efficient easily adaptable system for interpreting natural language queries}.
\newblock \emph{American Journal of Computational Linguistics}, 8(3-4):110--122.

\bibitem[{Xu et~al.(2017)Xu, Liu, and Song}]{xu2017sqlnet}
Xiaojun Xu, Chang Liu, and Dawn Song. 2017.
\newblock \href {http://arxiv.org/abs/1711.04436} {Sqlnet: Generating structured queries from natural language without reinforcement learning}.

\bibitem[{Yin et~al.(2016)Yin, Lu, Li, and Ben}]{yin-etal-2016-neural}
Pengcheng Yin, Zhengdong Lu, Hang Li, and Kao Ben. 2016.
\newblock \href {https://doi.org/10.18653/v1/W16-0105} {Neural enquirer: Learning to query tables in natural language}.
\newblock In \emph{Proceedings of the Workshop on Human-Computer Question Answering}, pages 29--35, San Diego, California. Association for Computational Linguistics.

\bibitem[{Yu et~al.(2018)Yu, Zhang, Yang, Yasunaga, Wang, Li, Ma, Li, Yao, Roman, Zhang, and Radev}]{yu-etal-2018-spider}
Tao Yu, Rui Zhang, Kai Yang, Michihiro Yasunaga, Dongxu Wang, Zifan Li, James Ma, Irene Li, Qingning Yao, Shanelle Roman, Zilin Zhang, and Dragomir Radev. 2018.
\newblock \href {https://doi.org/10.18653/v1/D18-1425} {{S}pider: A large-scale human-labeled dataset for complex and cross-domain semantic parsing and text-to-{SQL} task}.
\newblock In \emph{Proceedings of the 2018 Conference on Empirical Methods in Natural Language Processing}, pages 3911--3921, Brussels, Belgium. Association for Computational Linguistics.

\bibitem[{Yu et~al.(2019)Yu, Zhang, Yang, Yasunaga, Wang, Li, Ma, Li, Yao, Roman, Zhang, and Radev}]{yu2019spider}
Tao Yu, Rui Zhang, Kai Yang, Michihiro Yasunaga, Dongxu Wang, Zifan Li, James Ma, Irene Li, Qingning Yao, Shanelle Roman, Zilin Zhang, and Dragomir Radev. 2019.
\newblock \href {http://arxiv.org/abs/1809.08887} {Spider: A large-scale human-labeled dataset for complex and cross-domain semantic parsing and text-to-sql task}.

\bibitem[{Yuan et~al.(2023)Yuan, Liu, Zi, Liu, Peng, and Lou}]{yuan2023evaluating}
Zhiqiang Yuan, Junwei Liu, Qiancheng Zi, Mingwei Liu, Xin Peng, and Yiling Lou. 2023.
\newblock \href {http://arxiv.org/abs/2308.01240} {Evaluating instruction-tuned large language models on code comprehension and generation}.

\bibitem[{Zhang et~al.(2023{\natexlab{a}})Zhang, Cao, Chen, Xu, and Yu}]{zhang2023actsql}
Hanchong Zhang, Ruisheng Cao, Lu~Chen, Hongshen Xu, and Kai Yu. 2023{\natexlab{a}}.
\newblock \href {http://arxiv.org/abs/2310.17342} {Act-sql: In-context learning for text-to-sql with automatically-generated chain-of-thought}.

\bibitem[{Zhang et~al.(2023{\natexlab{b}})Zhang, Li, Li, Li, and Jin}]{zhang-etal-2023-self}
Kechi Zhang, Zhuo Li, Jia Li, Ge~Li, and Zhi Jin. 2023{\natexlab{b}}.
\newblock \href {https://doi.org/10.18653/v1/2023.acl-long.45} {Self-edit: Fault-aware code editor for code generation}.
\newblock In \emph{Proceedings of the 61st Annual Meeting of the Association for Computational Linguistics (Volume 1: Long Papers)}, pages 769--787, Toronto, Canada. Association for Computational Linguistics.

\bibitem[{Zhong et~al.(2020)Zhong, Yu, and Klein}]{zhong-etal-2020-semantic}
Ruiqi Zhong, Tao Yu, and Dan Klein. 2020.
\newblock \href {https://doi.org/10.18653/v1/2020.emnlp-main.29} {Semantic evaluation for text-to-{SQL} with distilled test suites}.
\newblock In \emph{Proceedings of the 2020 Conference on Empirical Methods in Natural Language Processing (EMNLP)}, pages 396--411, Online. Association for Computational Linguistics.

\end{thebibliography}

\newpage

\appendix

\section{Analysis of Column Names in Spider and Termite}
\label{app:eq_abbr}
The following Table present the percentage of column names that consist in abbreviations or compound nouns on both Spider and \originalDS{} dataset. On average, both datasets presents a similar distributions of this kind of columns names.
The equivalence in terms of abbreviations and compound nouns, as discussed in Section \ref{sec:originalDS} is crucial to a fair evaluation during the estimation of data contamination (Section \ref{sec:dc_method}).

\begin{table}[h!]
\resizebox{\linewidth}{!}{
\begin{tabular}{l|l|l|l|}
\cline{2-4}
                                                  & Table                          & Compound & Abbreviation \\ \hline
\multicolumn{1}{|l|}{\multirow{10}{*}{Termite}} & bowling                        & 0.63     & 0.00         \\ \cline{2-4} 
\multicolumn{1}{|l|}{}                            & centri                         & 0.48     & 0.16         \\ \cline{2-4} 
\multicolumn{1}{|l|}{}                            & coronavirus                    & 0.55     & 0.15         \\ \cline{2-4} 
\multicolumn{1}{|l|}{}                            & farma                          & 0.62     & 0.29         \\ \cline{2-4} 
\multicolumn{1}{|l|}{}                            & farmacia                       & 0.65     & 0.15         \\ \cline{2-4} 
\multicolumn{1}{|l|}{}                            & galleria                       & 0.20     & 0.00         \\ \cline{2-4} 
\multicolumn{1}{|l|}{}                            & hackathon                      & 0.62     & 0.00         \\ \cline{2-4} 
\multicolumn{1}{|l|}{}                            & pratica                        & 0.33     & 0.00         \\ \cline{2-4} 
\multicolumn{1}{|l|}{}                            & recensioni                     & 0.56     & 0.00         \\ \cline{2-4} 
\multicolumn{1}{|l|}{}                            & voli                           & 0.48     & 0.43         \\ \hline
\multicolumn{1}{|l|}{\multirow{20}{*}{Spider}}    & battle\_death                  & 0.33     & 0.00         \\ \cline{2-4} 
\multicolumn{1}{|l|}{}                            & car\_1                         & 0.30     & 0.09         \\ \cline{2-4} 
\multicolumn{1}{|l|}{}                            & concert\_singer                & 0.48     & 0.00         \\ \cline{2-4} 
\multicolumn{1}{|l|}{}                            & course\_teach                  & 0.60     & 0.00         \\ \cline{2-4} 
\multicolumn{1}{|l|}{}                            & cre\_Doc\_Template\_Mgt        & 1.00     & 0.00         \\ \cline{2-4} 
\multicolumn{1}{|l|}{}                            & dog\_kennels                   & 0.80     & 0.04         \\ \cline{2-4} 
\multicolumn{1}{|l|}{}                            & employee\_hire\_evaluation     & 0.59     & 0.00         \\ \cline{2-4} 
\multicolumn{1}{|l|}{}                            & flight\_2                      & 0.54     & 0.23         \\ \cline{2-4} 
\multicolumn{1}{|l|}{}                            & museum\_visit                  & 0.67     & 0.17         \\ \cline{2-4} 
\multicolumn{1}{|l|}{}                            & network\_1                     & 0.57     & 0.00         \\ \cline{2-4} 
\multicolumn{1}{|l|}{}                            & orchestra                      & 0.57     & 0.00         \\ \cline{2-4} 
\multicolumn{1}{|l|}{}                            & pets\_1                        & 0.64     & 0.36         \\ \cline{2-4} 
\multicolumn{1}{|l|}{}                            & poker\_player                  & 0.64     & 0.00         \\ \cline{2-4} 
\multicolumn{1}{|l|}{}                            & real\_estate\_properties       & 0.95     & 0.41         \\ \cline{2-4} 
\multicolumn{1}{|l|}{}                            & singer                         & 0.60     & 0.00         \\ \cline{2-4} 
\multicolumn{1}{|l|}{}                            & student\_transcripts\_tracking & 0.93     & 0.04         \\ \cline{2-4} 
\multicolumn{1}{|l|}{}                            & tvshow                         & 0.52     & 0.04         \\ \cline{2-4} 
\multicolumn{1}{|l|}{}                            & voter\_1                       & 0.67     & 0.11         \\ \cline{2-4} 
\multicolumn{1}{|l|}{}                            & World\_1                       & 0.42     & 0.15         \\ \cline{2-4} 
\multicolumn{1}{|l|}{}                            & wta\_1                         & 0.86     & 0.28         \\ \hline
\end{tabular}
}
%\caption{The number of compound and abbreviation on the number of columns per database}
\end{table}

\section{Measuring Hardness of queries in Spider and \originalDS{}}
\label{app:dataset_equivalence_test}
As described in Section \ref{sec:dataset_equivalence_test}, we need to ensure that Spider and \originalDS{} are comparable in terms of hardness.
\originalDS{} is designed with a similar annotation protocol; however, a similarity in terms of the hardness of the natural language questions used is hard to quantify.
For this reason, we asked 10 SQL-proficient annotators to perform a simple yet effective test to measure how difficult it is for them to translate questions both from Spider and from \originalDS{}. The main idea is that if they can translate both Spider and \originalDS{} questions with the same level of accuracy, then it means that the level of challenge is similar on both datasets.

In particular, given an E-R database schema and a natural language utterance, each test question asks the annotator to choose from three options the SQL query that satisfies the request. All three of the options are syntactically correct SQL queries, but the incorrect answers are semantically different from the correct one.
The first incorrect option is designed by the authors, perturbing the correct answer by removing or replacing some operations or some retrieved columns, changing the field and tables names with non-matching ones.
The second incorrect answer is, instead, another query extracted from the same dataset as the correct one. The selected query is the most similar under the Bag of Words assumption with respect to the correct one. The similarity of two queries, in order to retrieve this third option, is measured via cosine similarity of their BOW vector representations.

The complete test is composed of 20 randomly selected queries from each dataset, Hence, the resulting 40 questions are shared to 10 SQL-proficient annotators: 60\% of them are Computer Science Master students, the remaining are already graduated. Five of the annotators work in a field that requires daily use of the SQL query language.
Finally, we further divided the test into two trials of 20 queries each and administered it to the annotators at two different times to limit the presence of errors due to gradual loss of concentration.

\section{Assessing the presence of Data Contamination}
\label{app:dc_db_results}
The following two tables show the DC-accuracy of GPT-3.5 on the Spider (Table \ref{table:app_dc_spider}) and \originalDS{} (Table \ref{table:app_dc_termite}). Notice that, as discussed in Section \ref{sec:dc_results}, the overall performance in terms of DC accuracy on the Spider dataset is higher than that observed on the \originalDS{} dataset. Those results indicate the presence of data contamination.

\begin{table}[t!]
\begin{tabular}{|l|l|}
\hline
Database & DC-accuracy \\
\hline
battle\_death                  & 0.16 \\ \hline
car\_1                         & 0.00 \\ \hline
concert\_singer                & 0.78 \\ \hline
course\_teach                  & 0.00 \\ \hline
cre\_Doc\_Template\_Mgt        & 0.40 \\ \hline
dog\_kennels                   & 0.52 \\ \hline
employee\_hire\_evaluation     & 0.20 \\ \hline
flight\_2                      & 0.00 \\ \hline
museum\_visit                  & 0.00 \\ \hline
network\_1                     & 1.00 \\ \hline
orchestra                      & 0.43 \\ \hline
pets\_1                        & 0.50 \\ \hline
poker\_player                  & 0.50 \\ \hline
real\_estate\_properties       & 0.46 \\ \hline
singer                         & 0.00 \\ \hline
student\_transcripts\_tracking & 0.22 \\ \hline
tvshow                         & 0.00 \\ \hline
voter\_1                       & 1.00 \\ \hline
wta\_1                         & 0.16 \\ \hline
\end{tabular}
\caption{GPT-3.5 DC-accuracy across the different databases in Spider}
\label{table:app_dc_spider}
\end{table}

\begin{table}[h!]
\begin{tabular}{|l|l|}
\hline
Database & DC-accuracy \\
\hline
bowling     & 0.14 \\ \hline
centri      & 0.00 \\ \hline
coronavirus & 0.44 \\ \hline
farma       & 0.00 \\ \hline
farmacia    & 0.00 \\ \hline
galleria    & 0.00 \\ \hline
hackathon   & 0.33 \\ \hline
pratica     & 0.00 \\ \hline
recensioni  & 0.40 \\ \hline
voli        & 0.00 \\ \hline
\end{tabular}
\caption{GPT-3.5 DC-accuracy across the different databases in \originalDS{}}
\label{table:app_dc_termite}
\end{table}

\newpage
\newpage
\newpage

\begin{table*}[h]
\section{Text-to-SQL GPT-3.5 detailed performances}
\label{app:t2seval_dbdetail}
The following Table shows the results for each database in the Text-to-SQL task both for Spider and \originalDS{} dataset. Notice that the accuracy decreases as the hardness increases and that on \originalDS{}, results are generally lower.

\resizebox{\linewidth}{!}{
    \begin{tabular}{ll|lllll|}
    \cline{3-7}
                                                          &            & \multicolumn{5}{l|}{Termite} \\ \cline{2-7} 
    \multicolumn{1}{l|}{}                                 & difficulty & \multicolumn{1}{l|}{hackathon} & \multicolumn{1}{l|}{galleria} & \multicolumn{1}{l|}{recensioni} & \multicolumn{1}{l|}{centri} & \multicolumn{1}{l|}{pratica} \\ \hline
    \multicolumn{1}{|l|}{\multirow{4}{*}{Original}} & easy       & \multicolumn{1}{l|}{60.0}      & \multicolumn{1}{l|}{80.0}     & \multicolumn{1}{l|}{40.0}       & \multicolumn{1}{l|}{100.0}  & \multicolumn{1}{l|}{60.0}    \\ \cline{2-7} 
    \multicolumn{1}{|l|}{}                                & medium     & \multicolumn{1}{l|}{50.0}      & \multicolumn{1}{l|}{100.0}    & \multicolumn{1}{l|}{40.0}       & \multicolumn{1}{l|}{100.0}  & \multicolumn{1}{l|}{50.0}    \\ \cline{2-7} 
    \multicolumn{1}{|l|}{}                                & hard       & \multicolumn{1}{l|}{25.0}      & \multicolumn{1}{l|}{66.66}    & \multicolumn{1}{l|}{0.0}        & \multicolumn{1}{l|}{0.0}    & \multicolumn{1}{l|}{33.33}   \\ \cline{2-7} 
    \multicolumn{1}{|l|}{}                                & extra      & \multicolumn{1}{l|}{50.0}      & \multicolumn{1}{l|}{30.0}     & \multicolumn{1}{l|}{0.0}        & \multicolumn{1}{l|}{25.0}   & \multicolumn{1}{l|}{57.14}   \\ \hline
    \multicolumn{1}{|l|}{\multirow{4}{*}{ATD}}            & easy       & \multicolumn{1}{l|}{60.0}      & \multicolumn{1}{l|}{40.0}     & \multicolumn{1}{l|}{40.0}       & \multicolumn{1}{l|}{60.0}   & \multicolumn{1}{l|}{60.0}    \\ \cline{2-7} 
    \multicolumn{1}{|l|}{}                                & medium     & \multicolumn{1}{l|}{50.0}      & \multicolumn{1}{l|}{80.0}     & \multicolumn{1}{l|}{60.0}       & \multicolumn{1}{l|}{71.42}  & \multicolumn{1}{l|}{50.0}    \\ \cline{2-7} 
    \multicolumn{1}{|l|}{}                                & hard       & \multicolumn{1}{l|}{25.0}      & \multicolumn{1}{l|}{0.0}      & \multicolumn{1}{l|}{50.0}       & \multicolumn{1}{l|}{0.0}    & \multicolumn{1}{l|}{33.33}   \\ \cline{2-7} 
    \multicolumn{1}{|l|}{}                                & extra      & \multicolumn{1}{l|}{33.33}     & \multicolumn{1}{l|}{30.0}     & \multicolumn{1}{l|}{0.0}        & \multicolumn{1}{l|}{25.0}   & \multicolumn{1}{l|}{57.14}   \\ \hline \\ 
    %new line
    \cline{3-7}
                                                          &            & \multicolumn{5}{l|}{Termite} \\ \cline{2-7} 
    \multicolumn{1}{l|}{}                                 & difficulty & \multicolumn{1}{l|}{coronavirus} & \multicolumn{1}{l|}{farmacia} & \multicolumn{1}{l|}{voli}  & \multicolumn{1}{l|}{bowling} & farma \\ \hline
    \multicolumn{1}{|l|}{\multirow{4}{*}{Original}} & easy       & \multicolumn{1}{l|}{60.0}        & \multicolumn{1}{l|}{60.0}     & \multicolumn{1}{l|}{80.0}  & \multicolumn{1}{l|}{100.0}   & 100.0 \\ \cline{2-7} 
    \multicolumn{1}{|l|}{}                                & medium     & \multicolumn{1}{l|}{60.0}        & \multicolumn{1}{l|}{40.0}     & \multicolumn{1}{l|}{100.0} & \multicolumn{1}{l|}{55.55}   & 75.0  \\ \cline{2-7} 
    \multicolumn{1}{|l|}{}                                & hard       & \multicolumn{1}{l|}{40.0}        & \multicolumn{1}{l|}{60.0}     & \multicolumn{1}{l|}{25.0}  & \multicolumn{1}{l|}{33.33}   & 0.0   \\ \cline{2-7} 
    \multicolumn{1}{|l|}{}                                & extra      & \multicolumn{1}{l|}{0.0}         & \multicolumn{1}{l|}{40.0}     & \multicolumn{1}{l|}{20.0}  & \multicolumn{1}{l|}{14.28}   & 75.0  \\ \hline
    \multicolumn{1}{|l|}{\multirow{4}{*}{ATD}}            & easy       & \multicolumn{1}{l|}{40.0}        & \multicolumn{1}{l|}{60.0}     & \multicolumn{1}{l|}{80.0}  & \multicolumn{1}{l|}{80.0}    & 100.0 \\ \cline{2-7} 
    \multicolumn{1}{|l|}{}                                & medium     & \multicolumn{1}{l|}{60.0}        & \multicolumn{1}{l|}{60.0}     & \multicolumn{1}{l|}{100.0} & \multicolumn{1}{l|}{55.55}   & 50.0  \\ \cline{2-7} 
    \multicolumn{1}{|l|}{}                                & hard       & \multicolumn{1}{l|}{0.0}         & \multicolumn{1}{l|}{60.0}     & \multicolumn{1}{l|}{25.0}  & \multicolumn{1}{l|}{33.33}   & 0.0   \\ \cline{2-7} 
    \multicolumn{1}{|l|}{}                                & extra      & \multicolumn{1}{l|}{40.0}        & \multicolumn{1}{l|}{20.0}     & \multicolumn{1}{l|}{20.0}  & \multicolumn{1}{l|}{14.28}   & 50.0  \\ \hline \\

% spider
\cline{3-7}
                                                &                     & \multicolumn{5}{l|}{Spider}                                                                                                                                                               \\ \cline{2-7} 
\multicolumn{1}{l|}{}                           & difficulty          & \multicolumn{1}{l|}{battle\_death} & \multicolumn{1}{l|}{car\_1}             & \multicolumn{1}{l|}{concert\_singer}    & \multicolumn{1}{l|}{course\_teach}     & cre\_Doc\_Template\_Mgt \\ \hline
\multicolumn{1}{|l|}{\multirow{4}{*}{Original}} & easy                & \multicolumn{1}{l|}{100.0}         & \multicolumn{1}{l|}{94.44}  & \multicolumn{1}{l|}{100.0}              & \multicolumn{1}{l|}{75.0}              & 100.0                   \\ \cline{2-7} 
\multicolumn{1}{|l|}{}                          & medium              & \multicolumn{1}{l|}{62.5}          & \multicolumn{1}{l|}{40.62}             & \multicolumn{1}{l|}{62.5}               & \multicolumn{1}{l|}{85.71} & 79.54       \\ \cline{2-7} 
\multicolumn{1}{|l|}{}                          & hard                & \multicolumn{1}{l|}{0.0}           & \multicolumn{1}{l|}{18.75}              & \multicolumn{1}{l|}{61.54}  & \multicolumn{1}{l|}{62.5}              & 40.0                    \\ \cline{2-7} 
\multicolumn{1}{|l|}{}                          & extra               & \multicolumn{1}{l|}{50.0}          & \multicolumn{1}{l|}{15.38} & \multicolumn{1}{l|}{50.0}               & \multicolumn{1}{l|}{}                  & 33.33       \\ \hline
\multicolumn{1}{|l|}{\multirow{4}{*}{ATD}}                          & easy   & \multicolumn{1}{l|}{100.0}         & \multicolumn{1}{l|}{88.89}  & \multicolumn{1}{l|}{100.0}              & \multicolumn{1}{l|}{75.0}              & 91.66       \\ \cline{2-7} 
\multicolumn{1}{|l|}{}                          & medium & \multicolumn{1}{l|}{62.5}          & \multicolumn{1}{l|}{50.0}               & \multicolumn{1}{l|}{66.66}  & \multicolumn{1}{l|}{85.71} & 79.54       \\ \cline{2-7} 
\multicolumn{1}{|l|}{}                          & hard   & \multicolumn{1}{l|}{0.0}           & \multicolumn{1}{l|}{18.75}              & \multicolumn{1}{l|}{76.92}  & \multicolumn{1}{l|}{62.5}              & 40.0                    \\ \cline{2-7} 
\multicolumn{1}{|l|}{}                          & extra  & \multicolumn{1}{l|}{25.0}          & \multicolumn{1}{l|}{385} & \multicolumn{1}{l|}{0.0}                & \multicolumn{1}{l|}{}                  & 50.0                    \\ \hline \\
% new line
\cline{3-7}
                                                &                     & \multicolumn{5}{l|}{Spider}                                                                                                                                                               \\ \cline{2-7} 
\multicolumn{1}{l|}{}                           & difficulty          & \multicolumn{1}{l|}{dog\_kennels}      & \multicolumn{1}{l|}{employee\_hire\_evaluation} & \multicolumn{1}{l|}{flight\_2}          & \multicolumn{1}{l|}{museum\_visit}     & \multicolumn{1}{l|}{network\_1}        \\ \hline
\multicolumn{1}{|l|}{\multirow{4}{*}{Original}} & easy                & \multicolumn{1}{l|}{90.0}              & \multicolumn{1}{l|}{100.0}                      & \multicolumn{1}{l|}{84.62}  & \multicolumn{1}{l|}{100.0}             & \multicolumn{1}{l|}{100.0}             \\ \cline{2-7} 
\multicolumn{1}{|l|}{}                          & medium              & \multicolumn{1}{l|}{80.55} & \multicolumn{1}{l|}{92.86}          & \multicolumn{1}{l|}{76.66}  & \multicolumn{1}{l|}{87.5}              & \multicolumn{1}{l|}{77.27} \\ \cline{2-7} 
\multicolumn{1}{|l|}{}                          & hard                & \multicolumn{1}{l|}{60.0}              & \multicolumn{1}{l|}{80.0}                       & \multicolumn{1}{l|}{62.5}               & \multicolumn{1}{l|}{66.66} & \multicolumn{1}{l|}{56.25}             \\ \cline{2-7} 
\multicolumn{1}{|l|}{}                          & extra               & \multicolumn{1}{l|}{53.85} & \multicolumn{1}{l|}{0.0}                        & \multicolumn{1}{l|}{12.5}               & \multicolumn{1}{l|}{25.0}              & \multicolumn{1}{l|}{83.33} \\ \hline
\multicolumn{1}{|l|}{\multirow{4}{*}{ATD}}                          & easy   & \multicolumn{1}{l|}{90.0}              & \multicolumn{1}{l|}{100.0}                      & \multicolumn{1}{l|}{92.31}   & \multicolumn{1}{l|}{100.0}             & \multicolumn{1}{l|}{100.0}             \\ \cline{2-7} 
\multicolumn{1}{|l|}{}                          & medium & \multicolumn{1}{l|}{77.78} & \multicolumn{1}{l|}{100.0}                      & \multicolumn{1}{l|}{46.66} & \multicolumn{1}{l|}{100.0}             & \multicolumn{1}{l|}{81.82} \\ \cline{2-7} 
\multicolumn{1}{|l|}{}                          & hard   & \multicolumn{1}{l|}{80.0}              & \multicolumn{1}{l|}{80.0}                       & \multicolumn{1}{l|}{50.0}               & \multicolumn{1}{l|}{100.0}             & \multicolumn{1}{l|}{62.5}              \\ \cline{2-7} 
\multicolumn{1}{|l|}{}                          & extra  & \multicolumn{1}{l|}{50.0}              & \multicolumn{1}{l|}{25.0}                       & \multicolumn{1}{l|}{6.25}               & \multicolumn{1}{l|}{0.0}               & \multicolumn{1}{l|}{50.0}              \\ \hline \\
% new line
\cline{3-7}
                                                &                     & \multicolumn{5}{l|}{Spider}                                                                                                                                                               \\ \cline{2-7} 
\multicolumn{1}{l|}{}                           & difficulty          & \multicolumn{1}{l|}{orchestra}         & \multicolumn{1}{l|}{pets\_1}           & \multicolumn{1}{l|}{poker\_player} & \multicolumn{1}{l|}{real\_estate\_properties} & \multicolumn{1}{l|}{singer}            \\ \hline
\multicolumn{1}{|l|}{\multirow{4}{*}{Original}} & easy                & \multicolumn{1}{l|}{85.71} & \multicolumn{1}{l|}{100.0}             & \multicolumn{1}{l|}{93.75}         & \multicolumn{1}{l|}{100.0}                    & \multicolumn{1}{l|}{100.0}             \\ \cline{2-7} 
\multicolumn{1}{|l|}{}                          & medium              & \multicolumn{1}{l|}{83.33} & \multicolumn{1}{l|}{81.82} & \multicolumn{1}{l|}{100.0}         & \multicolumn{1}{l|}{50.0}                     & \multicolumn{1}{l|}{100.0}             \\ \cline{2-7} 
\multicolumn{1}{|l|}{}                          & hard                & \multicolumn{1}{l|}{83.33} & \multicolumn{1}{l|}{50.0}              & \multicolumn{1}{l|}{62.5}          & \multicolumn{1}{l|}{0.0}                      & \multicolumn{1}{l|}{33.33} \\ \cline{2-7} 
\multicolumn{1}{|l|}{}                          & extra               & \multicolumn{1}{l|}{50.0}              & \multicolumn{1}{l|}{30.0}              & \multicolumn{1}{l|}{}              & \multicolumn{1}{l|}{}                         & \multicolumn{1}{l|}{}                  \\ \hline
\multicolumn{1}{|l|}{\multirow{4}{*}{ATD}}                          & easy   & \multicolumn{1}{l|}{100.0}             & \multicolumn{1}{l|}{100.0}             & \multicolumn{1}{l|}{87.5}          & \multicolumn{1}{l|}{100.0}                    & \multicolumn{1}{l|}{83.33} \\ \cline{2-7} 
\multicolumn{1}{|l|}{}                          & medium & \multicolumn{1}{l|}{77.78} & \multicolumn{1}{l|}{68.18} & \multicolumn{1}{l|}{100.0}         & \multicolumn{1}{l|}{0.0}                      & \multicolumn{1}{l|}{88.89} \\ \cline{2-7} 
\multicolumn{1}{|l|}{}                          & hard   & \multicolumn{1}{l|}{83.33} & \multicolumn{1}{l|}{66.66} & \multicolumn{1}{l|}{50.0}          & \multicolumn{1}{l|}{0.0}                      & \multicolumn{1}{l|}{33.33} \\ \cline{2-7} 
\multicolumn{1}{|l|}{}                          & extra  & \multicolumn{1}{l|}{50.0}              & \multicolumn{1}{l|}{30.0}              & \multicolumn{1}{l|}{}              & \multicolumn{1}{l|}{}                         & \multicolumn{1}{l|}{}                  \\ \hline \\
% new row
\cline{3-7}
                                                &                     & \multicolumn{5}{l|}{Spider}                                                                                                                                                               \\ \cline{2-7} 
\multicolumn{1}{l|}{}                           & difficulty          & \multicolumn{1}{l|}{student\_transcripts\_tracking} & \multicolumn{1}{l|}{tvshow}            & \multicolumn{1}{l|}{voter\_1}          & \multicolumn{1}{l|}{world\_1}           & wta\_1             \\ \hline
\multicolumn{1}{|l|}{\multirow{4}{*}{Original}} & easy                & \multicolumn{1}{l|}{65.38}              & \multicolumn{1}{l|}{80.0}              & \multicolumn{1}{l|}{66.66} & \multicolumn{1}{l|}{79.16}  & 87.5               \\ \cline{2-7} 
\multicolumn{1}{|l|}{}                          & medium              & \multicolumn{1}{l|}{62.5}                           & \multicolumn{1}{l|}{86.66} & \multicolumn{1}{l|}{100.0}             & \multicolumn{1}{l|}{67.39}   & 66.66  \\ \cline{2-7} 
\multicolumn{1}{|l|}{}                          & hard                & \multicolumn{1}{l|}{37.5}                           & \multicolumn{1}{l|}{60.0}              & \multicolumn{1}{l|}{}                  & \multicolumn{1}{l|}{50.0}               & 42.86 \\ \cline{2-7} 
\multicolumn{1}{|l|}{}                          & extra               & \multicolumn{1}{l|}{15.0}                           & \multicolumn{1}{l|}{0.0}               & \multicolumn{1}{l|}{50.0}              & \multicolumn{1}{l|}{26.66} & 0.0                \\ \hline
\multicolumn{1}{|l|}{\multirow{4}{*}{ATD}}                          & easy   & \multicolumn{1}{l|}{65.38}              & \multicolumn{1}{l|}{85.0}              & \multicolumn{1}{l|}{100.0}             & \multicolumn{1}{l|}{75.0}               & 87.5               \\ \cline{2-7} 
\multicolumn{1}{|l|}{}                          & medium & \multicolumn{1}{l|}{66.66}              & \multicolumn{1}{l|}{80.0}              & \multicolumn{1}{l|}{100.0}             & \multicolumn{1}{l|}{58.69}  & 63.33  \\ \cline{2-7} 
\multicolumn{1}{|l|}{}                          & hard   & \multicolumn{1}{l|}{25.0}                           & \multicolumn{1}{l|}{30.0}              & \multicolumn{1}{l|}{}                  & \multicolumn{1}{l|}{45.0}               & 21.43 \\ \cline{2-7} 
\multicolumn{1}{|l|}{}                          & extra  & \multicolumn{1}{l|}{25.0}                           & \multicolumn{1}{l|}{50.0}              & \multicolumn{1}{l|}{25.0}              & \multicolumn{1}{l|}{23.33} & 50.0               \\ \hline
                                                
\end{tabular}
}
\caption{Test-Suite Evaluation results for GPT-3.5}
\end{table*}

%%% statistiche
\newpage

\end{document}